\documentclass{article}

\usepackage{PRIMEarxiv}

\usepackage[utf8]{inputenc} 
\usepackage[T1]{fontenc}    
\usepackage{hyperref}       
\usepackage{url}            
\usepackage{booktabs}       
\usepackage{amsfonts}       
\usepackage{nicefrac}       
\usepackage{microtype}      
\usepackage{lipsum}
\usepackage{fancyhdr}       
\usepackage{graphicx}       
\usepackage{amsmath,amsfonts}
\usepackage{algorithmic}
\usepackage{algorithm}
\usepackage{array}
\usepackage{cuted}
\usepackage{textcomp}
\usepackage{stfloats}
\usepackage{verbatim}
\usepackage{epsfig}
\usepackage{amssymb}
\usepackage{subcaption}
\usepackage{multirow}

\usepackage{pifont}
\usepackage{float}
\usepackage{breqn}
\usepackage[normalem]{ulem} 

\usepackage{authblk}

\title{VB-Mitigator: An Open-source Framework for Evaluating and Advancing Visual Bias Mitigation}

\author[1,2]{Ioannis Sarridis} 
\author[1]{Christos Koutlis}
\author[1]{Symeon Papadopoulos}
\author[2]{Christos Diou}
\affil[1]{Information Technologies Institute, CERTH, Greece}
\affil[2]{Department of Informatics and Telematics, Harokopio University of Athens, Greece}
\affil[ ]{\texttt{\{gsarridis,ckoutlis,papadop\}@iti.gr, \{isarridis,cdiou\}@hua.gr}}

\begin{document}
\maketitle

\begin{abstract}
Bias in computer vision models remains a significant challenge, often resulting in unfair, unreliable, and non-generalizable AI systems. Although research into bias mitigation has intensified, progress continues to be hindered by fragmented implementations and inconsistent evaluation practices. Disparate datasets and metrics used across studies complicate reproducibility, making it difficult to fairly assess and compare the effectiveness of various approaches. To overcome these limitations, we introduce the Visual Bias Mitigator (VB-Mitigator), an open-source framework designed to streamline the development, evaluation, and comparative analysis of visual bias mitigation techniques. VB-Mitigator offers a unified research environment encompassing 12 established mitigation methods, 7 diverse benchmark datasets. A key strength of VB-Mitigator is its extensibility, allowing for seamless integration of additional methods, datasets, metrics, and models. VB-Mitigator aims to accelerate research toward fairness-aware computer vision models by serving as a foundational codebase for the research community to develop and assess their approaches. To this end, we also recommend best evaluation practices and provide a comprehensive performance comparison among state-of-the-art methodologies.
\end{abstract}



\section{Introduction}
Computer vision (CV) systems have experienced significant growth and adoption across various fields \cite{dosovitskiy2020image, chen2020simple, wang2021generative}. These advancements have substantially improved automation, efficiency, and accuracy in numerous applications. However, the widespread presence of biases in CV models remains a critical and open challenge \cite{mehrabi2021survey, fabbrizzi2022survey, deandres2024frcsyn,ntoutsi2020bias, sarridis2024facex}. These biases, often arising from imbalanced training datasets, cause models to learn spurious correlations rather than meaningful, generalizable patterns \cite{ye2024spurious, barbano2022fairkl, li2023whac}. Consequently, such models frequently produce unreliable predictions, reduced generalizability, and outcomes that perpetuate Artificial Intelligence (AI) bias and stereotypes. For instance, facial recognition systems trained on skewed demographic distributions have exhibited racial biases, leading to harmful real-world consequences \cite{melzi2024frcsyn, sarridis2023towards}.

Although researchers have increasingly recognized these issues and dedicated significant effort toward bias mitigation strategies, the field suffers from fragmentation in implementation and evaluation practices, making it challenging to fairly assess and compare the efficacy of different mitigation approaches. Overall, this fragmentation complicates reproducibility, slows the development of robust solutions, and limits the community’s capacity to advance fairness-aware technologies efficiently.

To address these challenges, we introduce the Visual Bias Mitigator (VB-Mitigator), an open-source \footnote{\url{https://github.com/mever-team/vb-mitigator}} framework specifically created to facilitate standardized development, evaluation, and comparative analysis of visual bias mitigation methods. VB-Mitigator provides a cohesive research environment currently supporting 12 well established bias mitigation approaches, 7 commonly used datasets — including synthetic datasets, datasets involving standard protected attributes (such as gender, age, or race), background-related biases, and general-purpose CV datasets — and evaluation metrics to comprehensively assess fairness and robustness.
A key advantage of VB-Mitigator is its extensibility. The modular architecture facilitates effortless integration of new methods, datasets, evaluation metrics, and models. 

The primary aim of VB-Mitigator is to facilitate the development of fairer and more reliable CV systems by providing a comprehensive and extensible code-base. Furthermore, in the context of this work, we provide an extensive comparative evaluation of the bias mitigation methods included in VB-Mitigator, employing a unified and standardized evaluation protocol.

The main contributions of this work are the following:
\begin{itemize}
    \item An open-source framework designed to standardize and simplify the development, evaluation, and comparison of visual bias mitigation methods.
    \item A modular and extensible architecture that allows for the easy integration of new bias mitigation methods, datasets, evaluation metrics, and models.
    \item An extensive comparative evaluation of 12 established bias mitigation methods using VB-Mitigator. 
\end{itemize}
In the following sections, we detail the architecture and capabilities of VB-Mitigator, discuss the implemented bias mitigation approaches, datasets, and evaluation metrics, and present comprehensive experimental evaluations.

\section{Framework Architecture}

The VB-Mitigator is meticulously crafted to serve as a robust and adaptable platform for advancing research in visual bias mitigation. Its architectural design prioritizes modularity, extensibility, and reproducibility, allowing for effortless development and assessment bias mitigation methodologies. This section delves into the intricate structure and fundamental properties that empower VB-Mitigator.

\subsection{Core Components}
Building a comprehensive codebase for visual bias mitigation presents significant challenges. Existing techniques vary widely, intervening at diverse stages of the training pipeline—from data manipulation within data loaders and adjustments to loss functions, to the integration of external bias detection models and complex multi-stage training protocols. Compounding this, metric implementations are often dataset-specific, and limited to single or dual bias scenarios, hindering the creation of a generalizable evaluation platform. VB-Mitigator directly tackles these obstacles through an abstract and modular architecture built on PyTorch\footnote{\href{https://pytorch.org}{https://pytorch.org}}, chosen for its flexibility and robust ecosystem. By providing standardized interfaces for datasets, models, mitigation strategies, and evaluation metrics, the framework enables seamless integration and comparison of diverse approaches. 

\subsubsection{Datasets}
The dataset component (\texttt{datasets/}) encapsulates PyTorch Dataset classes, engineered to return dictionaries containing input images, targets, biases (or protected attributes), and sample indices via the \texttt{\_\_getitem\_\_} method. Furthermore, a \texttt{builder.py} module facilitates dataset construction, generating a comprehensive dictionary that includes critical metadata such as the number of classes, a list of protected attributes, the number of subgroups, class names, data subsets, and initial dataloaders. This metadata is essential for model initialization, metric computation, training orchestration, and dynamic dataloader updates.
\subsubsection{Mitigators}

\begin{figure}[t]
    \centering
    \includegraphics[width=\linewidth,trim={0cm 0cm 1.2cm 0cm},clip]{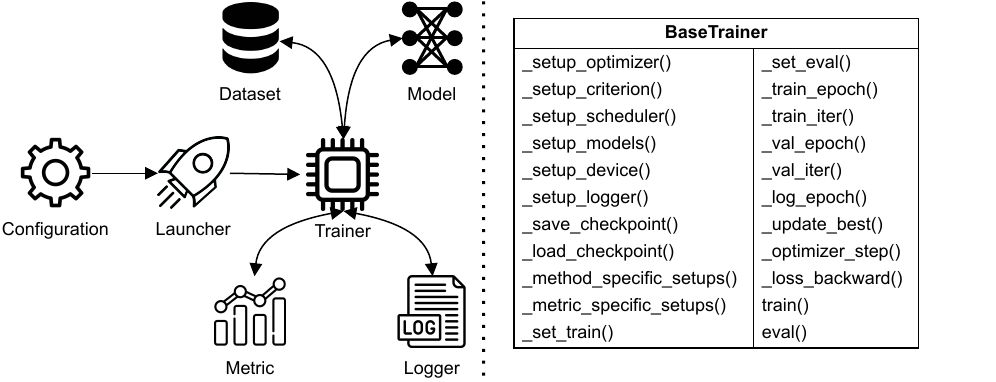}
    \caption{VB-Mitigator architecture revolves around the central Trainer component. The BaseTrainer within Trainer abstracts all training pipeline steps (right), facilitating the integration of new bias mitigation methods, by re-implementing only the functions of interest. }
    \label{fig:framework}
\end{figure}

The mitigator component (\texttt{mitigators/}) in VB-Mitigator acts as the core algorithmic engine, offering a flexible and standardized platform for bias mitigation algorithms. The \texttt{BaseTrainer} class establishes a comprehensive foundation for method implementation, defining functions for every stage of the training pipeline, including dataset handling, model training, metric computation, and logging (Figure~\ref{fig:framework}). Additionally, the framework supports method-specific configurations, facilitating the inclusion of preprocessing steps such as bias pseudo-label generation. This modular architecture allows each bias mitigation strategy to implement only the pipeline components where it actively intervenes, ensuring ease of integration and maintainability. 
\subsubsection{Models}
The models component (\texttt{models/}) contains a diverse collection of neural network architectures commonly used in visual bias mitigation research. It supports a range of architectures, including lightweight Convolutional Neural Networks (CNNs) designed for small-scale datasets such as Biased-MNIST\cite{bahng2020rebias}, widely adopted CNN  architectures like ResNets \cite{he2016deep} and EfficientNets \cite{tan2019efficientnet}, and modern vision transformers \cite{dosovitskiy2020image, liu2021swin} for more complex tasks. Additionally, it accommodates custom architectures tailored to specific bias mitigation methods, ensuring flexibility for diverse experimental setups.

\subsubsection{Metrics}
The metric component (\texttt{metrics/}) provides a comprehensive suite of evaluation metrics, tailored to assess fairness. Recognizing the common practice of employing multiple metrics for fairness evaluation (e.g., worst-group accuracy alongside average group accuracy), our implementation supports metric classes that encompass multiple measurements. Each metric class is further configured with two attributes: (i) an indicator specifying whether the metric is error-based or higher-is-better, and (ii) a designation of the primary evaluation metric that should be used for checkpoint selection.
\subsubsection{Tools and Utilities}

This component (\texttt{tools/}) encompasses essential utilities for experiment management, ensuring a streamlined workflow within the framework. It includes launcher scripts for experiment execution and critical functionalities such as logging and model checkpointing. The logging system supports both Wandb\footnote{\href{https://wandb.ai/site}{https://wandb.ai/site}} and TensorBoard\footnote{\href{https://www.tensorflow.org/tensorboard}{https://www.tensorflow.org/tensorboard}} for real-time monitoring while also generating detailed logs in human-readable format and structured CSV files that can be used for visualization purposes. Additionally, it manages checkpointing, storing model states at various stages, including the latest epoch, the best-performing model, and intermediate checkpoints, enabling experiment resuming.

\subsubsection{Configuration and Execution Scripts}
The configuration files act as a central control mechanism, allowing researchers to seamlessly switch between datasets, bias mitigation methods, and hyperparameters. Given the extensive number of configurable variables, we utilize the YACS\footnote{\href{https://github.com/rbgirshick/yacs}{https://github.com/rbgirshick/yacs}} library, which provides a structured and efficient way to manage configurations. This approach enhances flexibility while maintaining clarity in experimental setups.
Finally, the \texttt{scripts/} directory contains a collection of shell scripts designed to simplify and automate the execution of experiments across different bias mitigation methods and datasets.

\subsection{Framework Properties}
The design of VB-Mitigator emphasizes several key properties that enhance its utility as a comprehensive bias mitigation framework:
\begin{itemize}

    \item \textbf{Modularity:} The introduced framework adopts a modular architecture, where each component (i.e., datasets, models, mitigators, evaluation metrics, and logging) is encapsulated in separate modules, which allows for experimentation with different configurations without altering the core functionality of the system.

    \item \textbf{Extensibility:} VB-Mitigator is designed to facilitate seamless integration of new bias mitigation techniques, datasets, and evaluation metrics. As previously discussed, the abstract class definitions allow for introducing new methods with minimal effort, as they only need to define the pipeline components where their approach intervenes.

    \item \textbf{Reproducibility:} Ensuring consistency and reproducibility in experiments is a fundamental objective of VB-Mitigator. To achieve this, all operations involving stochasticity are explicitly seeded, while CUDA\footnote{\href{https://developer.nvidia.com/cuda-toolkit}{https://developer.nvidia.com/cuda-toolkit}} algorithms are configured to operate in a deterministic mode. However, complete determinism cannot be guaranteed due to variations in hardware and CUDA versions, which may introduce minor discrepancies in execution across different computational environments.

\end{itemize}

\section{Methodologies}

\subsection{Pleliminary Notation}

Here, we introduce the notation used throughout the paper to ensure consistency across all methods. Let $f(\mathbf{x}; \boldsymbol{\theta})$ denote a neural network parameterized by $\boldsymbol{\theta}$, which maps an input sample $\mathbf{x} \in \mathcal{X}$ to an output prediction $\hat{y} \in \mathcal{Y}$. The dataset consists of $N$ training samples, where each sample is associated with a target label $y \in \mathcal{Y}$ and may also include a bias attribute $a \in \mathcal{A}$ that introduces spurious correlations. The learned feature representation of an input is denoted as $\mathbf{z} = h(\mathbf{x}; \boldsymbol{\theta})$, where $h$ represents the feature extractor component of the model. The objective of a vanilla model is to learn the model parameters $\boldsymbol{\theta}$ that minimize the average loss over the training samples, formed as:
\[
\min_{\boldsymbol{\theta}} \frac{1}{N} \sum_{i=1}^{N} L(f(\mathbf{x}_i; \boldsymbol{\theta}), y_i).
\]

\subsection{Method Descriptions}
Bias mitigation methodologies can be broadly categorized based on their reliance on explicit bias annotations. Specifically, we distinguish between bias label unaware (BLU) methods, which operate without access to information about attributes inducing spurious correlations, and bias label aware (BLA) methods, which leverage such annotations. While BLA techniques often demonstrate superior performance in controlled settings, BLU approaches exhibit broader applicability, rendering them more suitable for real-world scenarios where bias annotations may be unavailable or unreliable. VB-Mitigator encompasses several approaches of both categories, featuring the following BLA methods: GroupDro \cite{sagawa2019distributionally}, Domain Independent (DI) \cite{wang2020DI}, Entangling and Disentangling (EnD) \cite{tartaglione2021end}, Bias Balance (BB) \cite{hong2021bb}, and Bias Addition (BAdd) \cite{sarridis2024badd}, as well as the following BLU methods: Learning from Failure (LfF) \cite{nam2020LfF}, Spectral Decouple (SD) \cite{pezeshki2021gradient},  Just Train Twice (JTT) \cite{liu2021just}, SoftCon \cite{hong2021bb}, Debiasing Alternate Networks (Debian) \cite{li2022discover}, Fairness Aware Representation Learning (FLAC and FLAC-B) \cite{sarridis2023flac}, and Mitigate Any Visual Bias (MAVias) \cite{sarridis2024mavias}. The key characteristics of these methods are reported in Table~\ref{tab:methods}. Below we briefly present the considered approaches.
\begin{table}[ht]
\centering
\caption{Summary of the integrated bias mitigation methods.}
\resizebox{\textwidth}{!}{%

\begin{tabular}{lccc}
\toprule
{Method} & {Bias Labels} & {Bias-Capturing Model} & {Summary} \\
\midrule
GroupDRO \cite{sagawa2019distributionally} & \checkmark & $\times$ & Minimizes worst-case loss across predefined groups. \\
DI \cite{wang2020DI}& \checkmark & \checkmark & Uses domain-specific classification heads for domain invariance. \\
EnD \cite{tartaglione2021end} & \checkmark & $\times$ & Disentangles bias representations and entangles class representations. \\
BB \cite{hong2021bb}& \checkmark & $\times$ & Balances bias in the logit space using prior bias information. \\
BAdd \cite{sarridis2024badd}& \checkmark & \checkmark & Adds bias-capturing features to training to discourage their use. \\
LfF \cite{nam2020LfF}& $\times$ & \checkmark & Reweights samples based on bias-conflicting predictions from an auxiliary model. \\
SD \cite{pezeshki2021gradient}& $\times$ & $\times$ & Regularizes network logits for spectral decoupling and bias robustness. \\
JTT \cite{liu2021just}& $\times$ & \checkmark & Reweights misclassified samples, assuming they are bias-conflicting. \\
SoftCon \cite{hong2021bb}& $\times$ & \checkmark & Uses a weighted contrastive loss based on bias-capturing features. \\
Debian \cite{li2022discover}& $\times$ & \checkmark & Alternates training of main and auxiliary models for debiasing. \\
FLAC \cite{sarridis2023flac}& $\times$ & \checkmark & Minimizes mutual information between representations and bias-capturing features. \\
MAVias \cite{sarridis2024mavias}& $\times$ & \checkmark & Infers and mitigates visual biases using foundation models and regularization. \\
\bottomrule
\end{tabular}}
\label{tab:methods}
\end{table}

\paragraph{\textbf{Group Distributionally Robust Optimization (GroupDro).}}
        GroupDro addresses bias by minimizing the worst-case loss across predefined groups, ensuring robustness to group-level biases. Groups are defined as combinations of $\mathcal{Y}$ and $\mathcal{A}$. The objective is to minimize the maximum loss across groups, defined as:
        \[
\min_{\boldsymbol{\theta}} \max_{g \in \mathcal{G}} \frac{1}{N_g} \sum_{i:g_i=g} L(f(\mathbf{x}_i; \boldsymbol{\theta}), y_i)
\]
        where $\mathcal{G}$ is the set of groups, $N_g$ is the number of samples in group $g$, and $g_i$ is the group assignment of sample $i$.

\paragraph{\textbf{Domain Independent (DI).}}
Domain Independent (DI) aims to mitigate bias by learning representations that are invariant across different domains. Unlike traditional methods, DI employs multiple classification heads, each corresponding to a distinct domain. For each input sample $\mathbf{x}$, belonging to domain $\alpha$, the model selects and utilizes only the logits produced by the classification head associated with domain $\alpha$. Let $f_\alpha(\mathbf{x}; \boldsymbol{\theta})$ denote the output logits from the classification head corresponding to domain $\alpha$, where $\boldsymbol{\theta}$ represents the model parameters.
The objective is to minimize the domain-specific loss while encouraging domain-invariant representations. This can be expressed as:
\[
\min_{\boldsymbol{\theta}} \left[ \frac{1}{N} \sum_{i=1}^{N} L(f_{\alpha_i}(\mathbf{x}_i; \boldsymbol{\theta}), y_i) \right].
\]
\paragraph{\textbf{Entangling and Disentangling (EnD).}}
        EnD explicitly disentangles representations of samples sharing the same bias label and entangles representations of samples under the same class with different bias labels. The loss function is:
       \[
\min_{\boldsymbol{\theta}} \Big[ L(f(\mathbf{x}; \boldsymbol{\theta}), y) + \lambda_{\text{EnD},1} L_{dis}(\mathbf{z}, \alpha) 
+ \lambda_{\text{EnD},2} L_{ent}(\mathbf{z}, \alpha, y) \Big]
\]
        where $L_{dis}$ is the disentanglement loss, $L_{ent}$ is the entanglement loss, and the regularization terms are denoted as $\lambda_{\text{EnD},1}$ and $\lambda_{\text{EnD},2}$.
        
\paragraph{\textbf{Bias Balance (BB).}}
        BB infers the unbiased distribution from the skewed one and it uses a loss function that balances the impact of biases in the logit space, expressed as:
        \[
\min_{\boldsymbol{\theta}} L(f(\mathbf{x}; \boldsymbol{\theta}) + \mathbf{p}, y)
\]
where $\mathbf{p}$ denotes the prior related to bias.

\paragraph{\textbf{Bias Addition (BAdd).}}
        BAdd explicitly adds bias to the training procedure and encourages the model not to learn features related to that bias. The objective has the following form:
        \[
\min_{\boldsymbol{\theta}} L(f(\mathbf{x}; \boldsymbol{\theta}), \mathbf{b}, y)
\]
where $\mathbf{b}$ denotes the bias features derived by a bias-capturing model.

\paragraph{\textbf{Learning from Failure (LfF).}}
        LfF employs a dual-model training paradigm, simultaneously training a main classification model and an auxiliary bias prediction model. The auxiliary model is trained to predict the biased attribute $\mathcal{A}$. By comparing the losses of both models within each mini-batch, LfF identifies bias-conflicting samples. These samples are subsequently re-weighted, adjusting their impact on the primary model's training. The optimization objective is defined as:
        \[
\min_{\boldsymbol{\theta}} \frac{1}{N} \sum_{i=1}^{N} w_i L(f(\mathbf{x}_i; \boldsymbol{\theta}), y_i).
\]
        where $w_i$ denotes the weight assigned sample $i$.

\paragraph{\textbf{Spectral Decouple (SD).}}
        SD shows that adding a regularization term to the network logits leads to a spectral decouple that enhances the network robustness on spurious correlations. The objective can be defined as:
        \[
\min_{\boldsymbol{\theta}} L(f(\mathbf{x}; \boldsymbol{\theta}), y) + \lambda_{\text{SD}} \|\hat{y}\|^2
\]
        where $\lambda_{\text{SD}}$ is the regularization weight. In contrast to other BLU methods, SD employs a generic approach to bias mitigation, operating without aiming at any bias-specific inference.
\paragraph{Just Train Twice (JTT).} JTT focuses on learning from misclassified examples, as they tend to be bias-conflicting samples. It uses a re-weighting scheme to prioritize these examples during training by modifying the dataloaders accordingly. 
\paragraph{Soft Contrastive (SoftCon).}
SoftCon is a weighted SupCon \cite{khosla2020supervised} loss that encourages feature similarity between samples with the same target label, weighted by the cosine distance between their feature derived by a bias-capturing model and representing the attribute introducing the spurious correlation. The weights can be expressed as:

\[
w_{i,j} = 1 - \frac{\mathbf{b}_i \mathbf{b}_j}{\|\mathbf{b}_i\| \|\mathbf{b}_j\|}.
\]

\paragraph{\textbf{Debiasing Alternate Networks (Debian).}}
Similarly to LfF, Debian uses an auxiliary model to encapsulate bias-related information. In particular, it employs a scheme that alternates between training a main network and an auxiliary network to mitigate bias. The predictions of the auxiliary model are used to assign weights to the main network's loss. Then, similar to LfF, the objective is:
\[
\min_{\boldsymbol{\theta}} \frac{1}{N} \sum_{i=1}^{N} w_i L(f(\mathbf{x}_i; \boldsymbol{\theta}), y_i).
\]

\paragraph{\textbf{Fairness Aware Representation Learning (FLAC).}}
FLAC focuses on learning fair representations by minimizing the dependence between features and sensitive attributes. The objective is to minimize the mutual information between representations and sensitive attributes, defined as:
\[
\min_{\boldsymbol{\theta}} L(f(\mathbf{x}; \boldsymbol{\theta}), y) + \lambda_{\text{FLAC}} I(\mathbf{z}, \alpha)
\]
where $\lambda_{\text{FLAC}}$ is a hyperparameter, $I(\mathbf{z}, \alpha)$ is the mutual information between representations and sensitive attributes, without accessing the $\mathcal{A}$ labels. To this end, FLAC employs a bias-capturing model trained to derive features $\mathbf{b}$. In scenarios where training a dedicated bias-capturing model is impractical (e.g., unknown biases), the biased vanilla model can be employed, resulting in the FLAC-B variant.  $\boldsymbol{\theta}$ represents the learnable parameters of the main model, and $\mathbf{z}$ and $\mathbf{a}$ are vector representations of the features and sensitive attributes, respectively.

\paragraph{\textbf{Mitigate Any Visual Bias (MAVias).}}
MAVias employs a two-stage approach. First, it infers potential visual biases by leveraging foundational models to generate descriptive tags for input images and assess their relevance to the target class. Subsequently, these potential biases are encoded using a vision-language model and incorporated into the training procedure as regularization, discouraging the model from learning spurious correlations. The minimization objective is defined as:
\[
\min_{\boldsymbol{\theta}, \boldsymbol{\phi}} L(f(\mathbf{x}; \boldsymbol{\theta}), y) + \lambda_{\text{MAVias},1} L_{reg}(\mathbf{x}, \mathbf{b}, \lambda_{\text{MAVias},2})
\]
where $\boldsymbol{\phi}$ represents the parameters of a projection layer that maps bias embeddings to the vision space, $L_{reg}$ is a regularization term, and $\lambda_{\text{MAVias},1}$ and $\lambda_{\text{MAVias},2}$ are hyperparameters. $\boldsymbol{\theta}$ represents the learnable parameters of the main model.

\section{Datasets}

VB-Mitigator supports diverse datasets to evaluate bias mitigation techniques across a spectrum of scenarios. These datasets encompass synthetic data with controlled biases, manually injected biases in established benchmarks, and general-purpose CV datasets, facilitating a comprehensive assessment of the compared methods.
\begin{table}[t]
\centering
\caption{Summary of the supported datasets.}
\resizebox{\textwidth}{!}{%
\begin{tabular}{lcccc}
\toprule
{Dataset}& {Data Type} & {Bias Type} & {\#Biases} & {Spurious Correlations} \\ \midrule
Biased-MNIST & digits & foreground color & 1 & 99\%-99.9\%\\
FB-Biased-MNIST & digits & foreground \& background color & 2 & 90\%-99\%\\
Biased-UTKFace & faces & demographics (race or age) & 1 & 90\%\\
Biased-CelebA & faces & demographics (gender) & 1 & 90\%\\
Waterbirds & bird species & background scene & 1 & 95\%\\
UrbanCars & car type & background scene \& co-occuring object & 2 & 95\%\\
ImageNet9 & general purpose & background \& texture & unknown & unknown\\
\bottomrule
\end{tabular}}
\label{tab:datasets}
\end{table}
\paragraph{\textbf{Biased-MNIST.}}
Biased-MNIST \cite{bahng2020rebias}, a modified version of the MNIST dataset, introduces background color correlations with digit labels. This dataset offers varying bias strengths, with 99\%, 99.5\%, 99.7\%, and 99.9\% co-occurrence levels commonly used to assess bias mitigation performance.
\paragraph{\textbf{FB-Biased MNIST.}}
Building upon Biased-MNIST, FB-Biased MNIST \cite{sarridis2024badd} introduces multiple biases by correlating both foreground and background colors with digit labels. This dataset, with suggested co-occurrence levels of 90\%, 95\%, and 99\%, provides a more challenging benchmark.
\paragraph{\textbf{Biased-UTKFace.}}
The UTKFace dataset \cite{zhifei2017cvpr}, comprising facial images with age, gender, and ethnicity annotations, serves as a foundation for Biased-UTKFace \cite{hong2021bb}. In this biased variant, gender is designated as the target variable, while race or age act as biasing attributes, exhibiting a 90\% co-occurrence with the target. This dataset is instrumental in examining demographic biases in facial attribute classification.
\paragraph{\textbf{Biased-CelebA.}}
Leveraging the large-scale CelebA dataset \cite{liu2015faceattributes}, which provides annotations for diverse facial attributes, Biased-CelebA \cite{hong2021bb} focuses on gender-related biases. Here, blonde hair or heavy makeup are the target attributes, with gender demonstrating a 90\% co-occurrence, enabling the study of attribute-specific gender biases.
\paragraph{\textbf{Waterbirds.}}
The Waterbirds dataset \cite{sagawa2019distributionally} facilitates the investigation of spurious correlations between bird species and their background environment. Featuring images of landbirds and waterbirds against corresponding terrestrial or aquatic backgrounds, it presents a 95\% co-occurrence between bird species and background, serving as a benchmark for evaluating context-dependent bias mitigation.
\paragraph{\textbf{UrbanCars.}}
UrbanCars \cite{li2023whac}, a dataset of car images, is designed to explore biases in object recognition. It examines biases in car type classification, where background (rural or urban) and co-occurring objects introduce biases with a 95\% co-occurrence rate with the target.
\paragraph{\textbf{ImageNet-9.}}
ImageNet-9 \cite{xiao2021noise}, a subset of ImageNet \cite{deng2009imagenet}, focuses on nine object categories. It is used to investigate background-related biases in large-scale object recognition, offering a more complex and realistic evaluation scenario where biases are unknown.

The progression of datasets used in visual bias mitigation reflects an increasing complexity, mirroring the evolution of research in this domain. Early studies predominantly focused on single-attribute biases, often utilizing synthetic or simplified datasets like Biased-MNIST. Recent research has shifted towards exploring multi-attribute biases, as exemplified by FB-Biased MNIST and UrbanCars, where many existing methods struggle to maintain performance. Furthermore, the inclusion of generic CV datasets like ImageNet-9 signifies a move towards addressing the challenges of real-world scenarios, where several interacting biases can occur. Table~\ref{tab:datasets} provides an overview of the considered datasets.

\section{Evaluation Metrics}
\label{sec:metrics}
Evaluating the efficacy of visual bias mitigation techniques necessitates careful consideration of appropriate performance metrics. While standard accuracy, defined as:
\[
\text{Acc} = \frac{|\{\mathbf{x} \in \mathcal{X} : \hat{y} = y\}|}{|\mathcal{X}|},
\]
remains a foundational measure, its utility is context-dependent.
Even when bias is uniformly distributed within a dataset, models may exhibit varying sensitivities to different bias attributes, rendering accuracy on a balanced test set (such as in Biased-MNIST and FB-Biased-MNIST) insufficient to capture the nuanced behavior of mitigation methods. On the other hand, in scenarios where biases are unknown and test sets are debiased through augmentations, such as by removing confounding background information in ImageNet9, accuracy constitutes a suitable measure.

Furthermore, Bias-Conflict Accuracy (BCA) is typically employed for Biased-CelebA and Biased-UTKFace datasets. BCA, defined as $\text{BCA} = \text{Acc}(\mathcal{X}_{BC})$ where $\mathcal{X}_{BC} = \{\mathbf{x} \in \mathcal{X} : a \text{ conflicts with } y\}$, attempt to focus on the underrepresented groups in the data the data. While this metric provides insights into performance on bias-conflicting samples, it cannot generalize to consider multiple, interacting biases.

Recognizing the limitations of these metrics, we advocate for the adoption of Worst Group Accuracy (WGA) and Average Accuracy (AvgAcc). WGA, defined as $\text{WGA} = \min_{g \in \mathcal{G}} \text{Acc}(g)$, and AvgAcc, defined as $\text{AvgAcc} = \frac{1}{|\mathcal{G}|} \sum_{g \in \mathcal{G}} \text{Acc}(g)$, offer a more robust evaluation framework, as they effectively capture performance disparities across subgroups, making them suitable for datasets with complex bias distributions and multiple bias attributes.

\section{Experiments}
This section presents a comparative analysis of the methods within VB-Mitigator, evaluated on Biased-CelebA, Waterbirds, UrbanCars, and ImageNet9. These datasets were selected to provide a representative evaluation across diverse bias scenarios, encompassing demographic, background scene, multi-attribute, and unknown biases, respectively.
\subsection{Evaluation Protocol}
Given the suitability of WGA and AvgAcc for datasets with explicitly defined biases, as outlined in Section~\ref{sec:metrics}, these metrics were employed for the evaluation of models on Biased-CelebA, Waterbirds, and UrbanCars. For ImageNet9, we utilized accuracy across its seven official test set variations, which facilitate a more thorough assessment of a model's dependence on non-target object features. These variations are:

\begin{itemize}
    \item \textbf{ORIGINAL}: The standard ImageNet9 test set, serving as the baseline.
    \item \textbf{ONLY-BG-B}: Images where only the background is visible, with the foreground object replaced by a black bounding box.
    \item \textbf{ONLY-BG-T}: Images where only the background is visible, with the foreground object replaced by an impainted bounding box.
    \item \textbf{NO-FG}: Images where the foreground object has been segmented and removed.
    \item \textbf{ONLY-FG}: Images where only the foreground object is visible, with a black background.
    \item \textbf{MIXED-RAND}: Images with the foreground object placed on a random background from a random class.
    \item \textbf{MIXED-NEXT}: Images with the foreground object placed on a random background from the next class in the dataset.
\end{itemize}
\begin{table}[htbp]
\centering
\caption{Worst group accuracy and average accuracy for CelebA, Waterbirds, and UrbanCars.}
\label{tab:resultsA}
\begin{tabular}{lcccccc}
\toprule
\multirow{2}{*}{Method} & \multicolumn{2}{c}{CelebA} & \multicolumn{2}{c}{Waterbirds} & \multicolumn{2}{c}{UrbanCars} \\
\cmidrule(lr){2-3} \cmidrule(lr){4-5} \cmidrule(lr){6-7}
 & WG Acc (\%) & Avg Acc (\%) & WG Acc (\%) & Avg Acc (\%) & WG Acc (\%) & Avg Acc (\%) \\
\midrule
GroupDRO \cite{sagawa2019distributionally} & 48.88\scriptsize{$\pm$7.6} & 80.76\scriptsize{$\pm$0.9} & 66.38\scriptsize{$\pm$2.2} & 82.80\scriptsize{$\pm$0.8} & 20.00\scriptsize{$\pm$14.2} & 57.76\scriptsize{$\pm$1.7} \\
DI \cite{wang2020DI} & 84.88\scriptsize{$\pm$1.8} & 91.16\scriptsize{$\pm$0.4} & 91.64\scriptsize{$\pm$0.7} & 94.14\scriptsize{$\pm$0.3} & \textbf{86.13}\scriptsize{$\pm$0.4} & \textbf{88.90}\scriptsize{$\pm$0.3} \\
EnD \cite{tartaglione2021end} & 54.56\scriptsize{$\pm$4.0} & 82.60\scriptsize{$\pm$0.6} & \textbf{94.72}\scriptsize{$\pm$0.5} & \textbf{96.10}\scriptsize{$\pm$0.2} & 47.52\scriptsize{$\pm$2.3} & 80.14\scriptsize{$\pm$0.5} \\
BB \cite{hong2021bb} & 85.68\scriptsize{$\pm$1.5} & 91.16\scriptsize{$\pm$0.4} & 93.24\scriptsize{$\pm$0.5} & 95.10\scriptsize{$\pm$0.2} & 66.24\scriptsize{$\pm$1.7} & 84.50\scriptsize{$\pm$1.3} \\
BAdd \cite{sarridis2024badd} & \textbf{88.98}\scriptsize{$\pm$2.1} & \textbf{91.50}\scriptsize{$\pm$0.4} & 94.00\scriptsize{$\pm$0.4} & 94.40\scriptsize{$\pm$0.3} & 84.64\scriptsize{$\pm$1.4} & 88.36\scriptsize{$\pm$1.5} \\
\midrule
LfF \cite{nam2020LfF} & 58.58\scriptsize{$\pm$6.7} & 82.50\scriptsize{$\pm$2.2} & 94.42\scriptsize{$\pm$0.3} & 96.10\scriptsize{$\pm$0.1} & 46.88\scriptsize{$\pm$2.1} & 80.30\scriptsize{$\pm$0.3} \\
SD \cite{pezeshki2021gradient} & 52.10\scriptsize{$\pm$3.5} & 82.50\scriptsize{$\pm$0.7} & 94.10\scriptsize{$\pm$0.4} & 96.24\scriptsize{$\pm$0.2} & 58.56\scriptsize{$\pm$2.7} & 83.54\scriptsize{$\pm$1.2} \\
JTT \cite{liu2021just} & 74.68\scriptsize{$\pm$0.6} & 81.54\scriptsize{$\pm$0.3} & 90.90\scriptsize{$\pm$0.8} & 94.10\scriptsize{$\pm$0.3} & 72.48\scriptsize{$\pm$4.1} & 81.76\scriptsize{$\pm$2.1} \\
SoftCon \cite{hong2021bb} & 34.34\scriptsize{$\pm$13.8} & 78.34\scriptsize{$\pm$1.8} & 47.04\scriptsize{$\pm$6.2} & 57.20\scriptsize{$\pm$2.1} & 34.72\scriptsize{$\pm$2.8} & 50.46\scriptsize{$\pm$2.0} \\
Debian \cite{li2022discover} & 49.22\scriptsize{$\pm$13.0} & 81.24\scriptsize{$\pm$3.2} & 94.74\scriptsize{$\pm$0.5} & 96.06\scriptsize{$\pm$0.4} & 48.96\scriptsize{$\pm$2.0} & 80.84\scriptsize{$\pm$0.7} \\
FLAC \cite{sarridis2023flac} & 84.32\scriptsize{$\pm$3.4} & 89.20\scriptsize{$\pm$0.9} & 91.08\scriptsize{$\pm$0.5} & 93.62\scriptsize{$\pm$0.6} & 62.56\scriptsize{$\pm$0.8} & 83.96\scriptsize{$\pm$0.5} \\
MAVias \cite{sarridis2024mavias} & \textbf{84.88}\scriptsize{$\pm$0.8} & \textbf{90.64}\scriptsize{$\pm$0.3} & \textbf{95.90}\scriptsize{$\pm$0.2} & \textbf{96.34}\scriptsize{$\pm$0.2} & \textbf{80.80}\scriptsize{$\pm$0.9} & \textbf{87.12}\scriptsize{$\pm$0.5} \\
\bottomrule
\end{tabular} 
\end{table}
\begin{table}[t]
\centering
\caption{BLU methods accuracy comparison across the 7 test sets of ImageNet9.} 
\label{tab:resultsB}
 \resizebox{\linewidth}{!}{
\begin{tabular}{lccccccc} 
\toprule
Method & MIXED-NEXT ($\uparrow$) & MIXED-RAND ($\uparrow$) 
& NO-FG ($\downarrow$) & ONLY-BG-B ($\downarrow$) & ONLY-BG-T ($\downarrow$) & ONLY-FG ($\uparrow$) & ORIGINAL ($\uparrow$)
\\ 
\midrule       

LfF \cite{nam2020LfF} & 78.70 \scriptsize{$\pm$0.1} & 81.47 \scriptsize{$\pm$0.2} 
& 61.07 \scriptsize{$\pm$0.1} & 34.82 \scriptsize{$\pm$0.2} & 44.46 \scriptsize{$\pm$0.0} & 88.99 \scriptsize{$\pm$0.2} & 94.34 \scriptsize{$\pm$0.2} \\
JTT \cite{liu2021just} & 84.43 \scriptsize{$\pm$0.1} & 86.16 \scriptsize{$\pm$0.5} 
& 61.09 \scriptsize{$\pm$2.0} & 32.04 \scriptsize{$\pm$1.0} & {36.62} \scriptsize{$\pm$4.7} & 92.09 \scriptsize{$\pm$0.5} & 97.71 \scriptsize{$\pm$0.1} \\
Debian \cite{li2022discover} & 83.02 \scriptsize{$\pm$0.4} & 85.64 \scriptsize{$\pm$0.3}
& 64.53 \scriptsize{$\pm$0.4}& 34.45 \scriptsize{$\pm$0.1}& 45.00 \scriptsize{$\pm$0.6}& 93.06 \scriptsize{$\pm$0.1}& {97.89} \scriptsize{$\pm$0.1}\\
SoftCon \cite{hong2021bb}& 28.15 \scriptsize{$\pm$2.20} & 30.53 \scriptsize{$\pm$3.17} & \textbf{28.02} \scriptsize{$\pm$3.05} &\textbf{ 19.85} \scriptsize{$\pm$2.36} & \textbf{24.07} \scriptsize{$\pm$2.46} & 33.00 \scriptsize{$\pm$5.63} & 47.47 \scriptsize{$\pm$4.92} \\
SD \cite{pezeshki2021gradient}& {87.56} \scriptsize{$\pm$0.57} & {88.92} \scriptsize{$\pm$0.74} & 62.60 \scriptsize{$\pm$1.05} & 31.42 \scriptsize{$\pm$2.93} & 40.81 \scriptsize{$\pm$3.00} & \textbf{93.71} \scriptsize{$\pm$0.71} & \textbf{98.16} \scriptsize{$\pm$0.06} \\
FLAC-B \cite{sarridis2023flac}& 84.60 \scriptsize{$\pm$0.46} & 86.62 \scriptsize{$\pm$0.45} &{59.84} \scriptsize{$\pm$1.67} & {29.71} \scriptsize{$\pm$0.53} & 40.38 \scriptsize{$\pm$1.28} & 92.72 \scriptsize{$\pm$0.73} & 97.89 \scriptsize{$\pm$0.09} \\					
MAVias \cite{sarridis2024mavias}& \textbf{88.26}\scriptsize{$\pm$0.1}  & \textbf{89.64}\scriptsize{$\pm$0.2}  
& {53.02}\scriptsize{$\pm$0.7} & {21.83}\scriptsize{$\pm$0.4} & {32.48}\scriptsize{$\pm$0.6} & 91.90\scriptsize{$\pm$0.4} & 96.92\scriptsize{$\pm$0.2} \\
\bottomrule
\end{tabular}}
\end{table}
\subsection{Implementation Details}

Below, we outline the data preprocessing, model architectures, and hyperparameters common to all methods, unless stated otherwise.

\paragraph{\textbf{Biased-CelebA.}}
We utilized the Biased-CelebA dataset with ``blond hair" as the target attribute and ``gender" as the spurious correlation. Images were resized to $224 \times 224$ pixels and normalized using ImageNet statistics. A ResNet18 architecture was employed, trained for 10 epochs with a batch size of 128. The Adam optimizer was used with an initial learning rate of $0.001$, which was reduced by a factor of $0.1$ at epochs 3 and 6. A weight decay of $0.0001$ was applied.

\paragraph{\textbf{Waterbirds.}}
Images were resized to $256 \times 256$ pixels, center-cropped to $224 \times 224$ pixels, and normalized using ImageNet statistics. A ResNet50 model was trained for 100 epochs with a batch size of 64. The Stochastic Gradient Descent (SGD) optimizer was used with a learning rate of $0.001$ and a weight decay of $0.0001$.

\paragraph{\textbf{UrbanCars.}}
Images were resized to $256 \times 256$ pixels, center-cropped to $224 \times 224$ pixels, and subjected to random rotations (up to 45 degrees) and horizontal flips. Normalization was performed using ImageNet statistics. A ResNet50 architecture was trained using SGD for 150 epochs with a batch size of 64 and a weight decay of $0.0001$.

\paragraph{\textbf{ImageNet9.}}
Images were resized to $256 \times 256$ pixels, center-cropped to $224 \times 224$ pixels, and normalized using ImageNet statistics. A ResNet50 model was trained for 30 epochs with a batch size of 64. The SGD optimizer was used with an initial learning rate of $0.001$, which was reduced by a factor of $0.5$ at epoch 25.

Method-specific hyperparameters were configured following the values recommended in their respective original publications. For GroupDro, a robust step size of 0.01 was used. In SoftCon, the Cross-Entropy loss was weighted by 0.01. The $\lambda_{\text{FLAC}}$ parameter was set to 30,000, 10,000, 10,000, and 100 for Biased-CelebA, Waterbirds, UrbanCars, and ImageNet9, respectively. For JTT, bias-conflicting samples were upweighted by a factor of 100, and a learning rate of 0.00001 with a weight decay of 1 was employed. For MAVias, the $(\lambda_{\text{MAVias},1}, \lambda_{\text{MAVias},2})$ parameters were set to (0.01, 0.5), (0.05, 0.6), (0.01, 0.4), and (0.001, 0.7) for Biased-CelebA, Waterbirds, UrbanCars, and ImageNet9, respectively. For SD, the $\lambda_{\text{SD}}$ parameter was set to 0.1. For EnD, the $\lambda_{\text{EnD},1}$ and $\lambda_{\text{EnD},2}$ parameters were both set to 1. Experiments were repeated for 5 random seeds on an NVIDIA A100 GPU.
\subsection{Results}
This section presents a comparative evaluation of bias mitigation methods across diverse datasets, encompassing varying bias scenarios. As shown in Table~\ref{tab:resultsA} BLA methods, such as DI and BAdd, generally exhibit superior WGA across all datasets, demonstrating the efficacy of leveraging explicit bias information. However, GroupDRO, EnD, and BB display performance variability, particularly on the UrbanCars dataset, where they all underperform. This stems from UrbanCars' multi-attribute bias structure, which contrasts with these methods' design for single-attribute biases. BLU methods, while typically achieving lower performance than BLA methods, reveal similar trends. Notably, MAVias demonstrates consistent performance across datasets, whereas SoftCon training is unstable, likely due to its strong dependence on the auxiliary model.

For ImageNet9, where biases are inherently unknown, only BLU methods are applicable. Note that here, we use the BLU variant of FLAC, denoted as FLAC-B, employing the vanilla model as the bias-capturing component. As shown in Table~\ref{tab:resultsB} SD and MAVias showcase robust generalization across the various test set configurations of ImageNet9. Conversely, SoftCon again fails to converge. 

\section{Limitations and Ethical Considerations}
While VB-Mitigator's architecture is designed to facilitate the seamless integration of existing visual bias mitigation approaches, it is important to acknowledge that future methodologies may introduce unforeseen integration challenges. Furthermore, it should be stressed that bias mitigation is an open research problem. The methods currently supported by VB-Mitigator, while effective, do not guarantee the creation of perfectly fair models. 

\section{Conclusion}
This paper introduced VB-Mitigator, an open-source framework designed to contribute to the critical challenge of bias in computer vision models. The fragmentation of implementation and evaluation practices hinders progress in bias mitigation. VB-Mitigator addresses this by providing a standardized platform, promoting reproducibility and fair comparisons. This framework serves as a foundational codebase for the research community, enabling the efficient development and assessment of fairness-aware approaches.
Future work will focus on expanding VB-Mitigator with new methods and datasets. A key direction is integrating methods that leverage foundation models for deriving potential biases \cite{sarridis2024mavias,ciranni2024say}. Such annotations will allow for evaluating fairness in a wide range of general-purpose computer vision datasets, where bias has not yet been explored, enhancing this way the framework's impact towards addressing bias in real-world scenarios.

\section*{Acknowledgments}
This research was supported by the EU Horizon Europe projects
MAMMOth (grant no. 101070285), ELIAS (grant no. 101120237), and ELLIOT (grant no. 101214398).



\bibliographystyle{unsrt}  
\bibliography{sample-ceur}

\end{document}